\documentclass[runningheads]{llncs}
\usepackage{graphicx}
\usepackage{amsmath,amssymb} 
\usepackage{color}
\usepackage{tabularx}
\usepackage{multirow}
\usepackage{arydshln}
\usepackage[nocompress]{cite}
\usepackage{float}
\usepackage{booktabs}
\usepackage{epsfig}
\usepackage{caption}

\newcommand{\keypoint}[1]
{\noindent\textbf{#1}\quad}
\usepackage[width=122mm,left=12mm,paperwidth=146mm,height=193mm,top=12mm,paperheight=217mm]{geometry}
\begin{document}
\pagestyle{headings}
\mainmatter

\title{Universal Sketch Perceptual Grouping} 

\titlerunning{Universal Sketch Perceptual Grouping}

\authorrunning{Ke Li et al.}

\author{Ke Li$^{1,2}$\quad
\and
Kaiyue Pang$^2$\quad
\and
Jifei Song$^2$\quad
\and
Yi-Zhe Song$^2$\quad
\and
Tao Xiang$^2$\quad
\and
Timothy M. Hospedales$^3$\quad
\and
Honggang Zhang$^1$\\
}


\institute{$^1$Beijing University of Posts and Telecommunications\\
$^2$SketchX, Queen Mary University of London, $^3$The University of Edinburgh\\
    \email{ \{like1990,zhhg\}@bupt.edu.cn, \{t.hospedales\}@ed.ac.uk}\\
    \email{ \{kaiyue.pang,j.song,yizhe.song,t.xiang\}@qmul.ac.uk}\\
}

\maketitle
\begin{figure}[H]
\centering
\includegraphics[width = 0.8\linewidth]{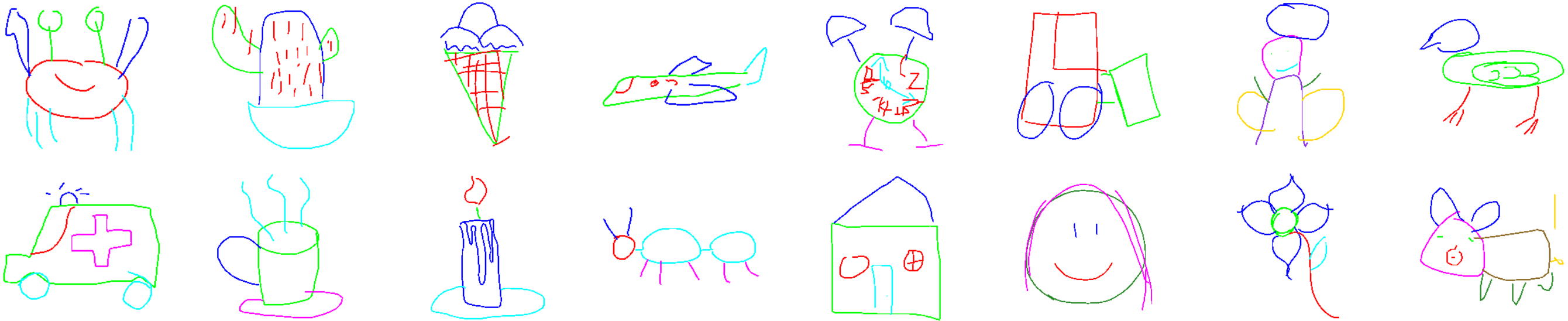}
\caption{Examples of the SPG dataset. Stroke groups are colour coded. }
\label{annotation}
\end{figure}
\begin{abstract}
In this work we aim to develop a universal sketch grouper.  That is, a grouper  that can be applied to sketches of any category in any domain to group constituent strokes/segments into semantically meaningful object parts.  The first obstacle to this goal is the lack of large-scale datasets with grouping annotation. To overcome this, we contribute  the largest sketch perceptual grouping (SPG) dataset to date, consisting of $20,000$ unique sketches evenly distributed over $25$ object categories. Furthermore, we propose a novel deep universal perceptual grouping model. The model is learned with both generative and discriminative losses.  The generative losses improve the generalisation ability of the model to unseen object categories and datasets. The discriminative losses include a local grouping loss and a novel global grouping loss to enforce global grouping consistency. We show that the proposed model significantly outperforms the state-of-the-art groupers. Further, we show that our grouper is useful for a number of sketch analysis tasks including sketch synthesis and fine-grained sketch-based image retrieval (FG-SBIR).
\end{abstract}

\section{Introduction}
Humans effortlessly detect objects and object parts out of a cluttered background. The Gestalt school of psychologists \cite{wagemans2012century1,wagemans2012century2} argued that this ability to perceptually group visual cues/patterns into objects is built upon a number of grouping principles, termed Gestalt laws of grouping. These include five categories, namely proximity, similarity, continuity, closure, and symmetry \cite{wertheimer1938laws}. 

Computer vision research in grouping or segmentation has long  exploited these perceptual grouping principles. For example, in image segmentation \cite{ren2003learning,Xia2017W,Wang2017Unsupervised, CP2016Deeplab}, pixel visual appearance similarity and local proximity are often used to group pixels into objects. These principles are exploited by the human visual system to robustly perform perceptual grouping in diverse contexts and for diverse object categories. Exploiting them is important for a universal grouping algorithm.

We aim to develop such a universal grouper for human free-hand sketches which takes a  sketch as input and groups the constituent strokes into semantic parts. Note that this is different from semantic segmentation for either photos \cite{CP2016Deeplab} or sketches \cite{sun2012free,huang2014data,schneider2016example}, where each segmented part is given a label, and the labels are often object category-dependent. Only the group relationship between strokes is predicted so that the grouper can \emph{universally} be applied to any object category. 


Very few existing studies \cite{qi2013sketching,Qi2015Making} investigate sketch perceptual grouping. These approaches compute hand-crafted features from each stroke and use the proximity and continuity principles to compute a stroke affinity matrix for subsequent clustering/grouping. They thus have a number of limitations: (i) Only two out of the five principles are exploited, while the unused ones such as closure are clearly useful in grouping human sketches which can be fragmented (see Fig.~\ref{annotation}). (ii) How the principles are  formulated is determined manually  rather than learned from data. (iii) Fixed weightings of different principles are used which are either manually set \cite{qi2013sketching} or learned \cite{Qi2015Making}. However, for different sketches, different principles could be used by humans with different weightings. Therefore a more dynamic sketch-specific grouping strategy is preferable. 
Nevertheless, the existing sketch perceptual grouping datasets \cite{sun2012free,Qi2015Making} are extremely small, containing 2,000 annotated sketches at most. This prevents more powerful and flexible deep neural network models from being developed.

The first contribution of this paper is to provide the first large-scale sketch perceptual grouping (SPG) dataset consisting of 20,000 sketches with ground truth grouping annotation, i.e., 10 times larger than the largest dataset to date \cite{Qi2015Making}. The sketches are collected from 25 representative object categories with 800 sketches per category. Some examples of the sketches and their annotation can be seen in  Fig.~\ref{annotation}. A dataset of such size makes the development of a deep universal grouper possible. 


Even with sufficient training samples, learning a deep universal sketch grouper is non-trivial. In particular, there are two challenges: how to make the deep grouper generalisable to unseen object categories  and domains without any training data from them; and how to design training losses that enforce both local (stroke pairwise) grouping consistency and global (whole sketch level) grouping consistency given variable number of strokes per sketch. Most losses used by existing deep models are for supervised classification tasks; grouping is closer to clustering than classification so few options exist. 

In this paper, we propose a novel deep sketch grouping model to overcome both challenges. Specifically, treating a sketch as a sequence of strokes/segments, our model is a sequence-to-sequence variational autoencoder (VAE). The reconstruction loss in this deep generative model forces the learned  representation to preserve information richer than required for the discriminative grouping task alone. This has been proven to be useful for improving model generalisation ability \cite{HinSal06}, critical for making the grouper universal. As for the discriminative grouping learning objectives, we deploy two losses: a pairwise stroke grouping loss enforcing local grouping consistency and a global grouping loss to enforce global grouping consistency. This separation of the local and global grouping losses enables us to balance the two and makes our model more robust against annotation noise. Based on the proposed grouper we develop a simple sketch synthesis model by grouping and abstracting photo edge maps. The synthesised sketches can be used to learn a strong \emph{unsupervised} fine-grained sketch-based image retrieval (FG-SBIR) model, i.e., using photos only.


Our contributions are as follows: (1) We contribute the largest sketch perceptual grouping dataset to date with extensive human annotation. To drive future research, we will make the dataset publicly available.
(2) For the first time, a deep universal sketch grouper is developed based on a novel deep sequence-to-sequence VAE with both generative and discriminative losses. (3)  Extensive experiments show the superiority of our grouper against existing ones, especially when evaluated on new categories or new domains. Its usefulness  on a number of sketch analysis tasks including sketch synthesis and FG-SBIR is also demonstrated.

\section{Related work}

\keypoint{Perceptual Grouping:} Humans can easily extract salient visual structure from apparent noise. Gestalt psychologists referred to this phenomenon as perceptual organisation \cite{wagemans2012century1,wagemans2012century2} and introduced the concept of perceptual grouping, which accounts for the observation that humans naturally group visual patterns into objects.  A set of simple Gestalt principles were further developed, including proximity,  similarity and continuity \cite{wertheimer1938laws}, with closure, connectedness and common fate introduced later, primarily for studying human vision systems \cite{amir1998generic,ren2003learning}. 

\keypoint{Sketch Groupers:} Very few studies exist on grouping sketch strokes into  parts. The most related studies are \cite{qi2013sketching,Qi2015Making}. They compute an affinity matrix between strokes using hand-crafted features based on  proximity and continuity principles. The two principles are combined with fixed weights learned from human annotated stroke groups. In contrast, we assume that when humans draw sketches and annotate them into groups, all grouping principles could be used. Importantly, using which ones and by how much are dependent on the specific sketch instance. Our model is thus a deep neural network that takes the sketch as input and aims to model all principles implicitly via both generative and discriminative grouping losses. It thus has the potential to perform principle selection and weighting dynamically according to the sketch input. We also provide a much larger dataset compared to the one provided in \cite{Qi2015Making}. We show that on both datasets, our model outperforms that in \cite{Qi2015Making} by a big margin. Note that perceptual grouping has been modelled using a deep autoencoder in \cite{LunACMTOG17}. However, the objective is to group discrete graphical patterns which has richer visual cues that make them more akin to the problem of image segmentation, and thus easier than grouping line drawings in sketches.

\keypoint{Sketch Semantic Segmentation:} A closely related problem to sketch grouping is sketch semantic  segmentation \cite{sun2012free,huang2014data,schneider2016example}\footnote{Their relationship is analogous to that between unsupervised image segmentation \cite{Xia2017W,Wang2017Unsupervised} and semantic segmentation \cite{CP2016Deeplab}.}. The key difference is that a sketch grouper is universal in that it can be applied to any object category as it only predicts whether strokes belong to the same group rather than what group. In contrast, sketch segmentation models need to predict the label of each group. As a result, typically one model is needed for each object category. Note that although two different problems are tackled, our work can potentially benefit sketch semantic segmentation in two ways: (i) The grouping principles modelled implicitly in our grouping model could be used for semantic segmentation, e.g., by modifying/fine-tuning our model to a fully supervised one. (ii) The SPG dataset also contains group ID labels for each category so can be used for developing deep segmentation models, which has not been possible to date due to the small sizes of existing sketch segmentation datasets \cite{sun2012free,huang2014data,schneider2016example}.

\keypoint{Sketch Stroke Analysis:} Like our model, a number of recent sketch models are based on stroke modelling. \cite{schneider2016example} studied stroke semantic segmentation. A sequence-to-sequence variational autoencoder is  used in \cite{ha2017neural} for a  different purpose of conditional sketch synthesis.  The work in \cite{Umar2018Abstraction} uses a sketch RNN for sketch abstraction problem by sequentially removing redundant strokes. A stroke-based model is naturally suited for perceptual grouping. Modelling Gestalt principles is harder if a sketch is treated as a 2D pixel array instead of strokes. 


\keypoint{Fine-grained SBIR:} FG-SBIR has been a recent focus in sketch analysis  \cite{li2014fine,yu2016sketch,song2016sketch,pang2017fgsbir,hu2018sketch,song2017sketch}. Training a FG-SBIR model typically requires expensive photo-sketch pair collection, which severely restricts its applicability to large number of object categories. In this work, we show that our universal grouper is general enough to be applied to edge maps computed from object photos. The edge maps can then be abstracted by removing the least important groups. The abstracted edge map can be used to substitute human sketches and form synthetic sketch-photo pairs for training a FG-SBIR model. We show that the performance of a model trained in this way approaches that of the same model trained with human labelled data, and is superior to the state-of-the-art unsupervised alternative \cite{Umar2018Abstraction}.

\section{Free-hand Sketch Perceptual Grouping Dataset}

We contribute the Sketch Perceptual Grouping dataset, the largest free-hand sketch perceptual grouping dataset to date. It contains $20,000$ sketches  distributed over $25$ categories with each sketch manually annotated into parts.

\keypoint{Category Selection} The sketches come from the QuickDraw dataset \cite{ha2017neural}, which is by far the largest sketch dataset. It contains $345$  categories of everyday objects. Out of these, $25$ are selected for SPG (see Table \ref{tab:mainresult}) based on following criteria:  (i) Complexity: the category should contain at least three semantic parts, meaning categories such as cloud and moon are out.  (ii) {Variety}: The selected categories need to be sufficiently different from each other to be appropriate for testing the grouper's generalisation ability to unseen classes. For example, only one of the four-legged animal classes is chosen. 

\begin{figure}[t]
\centering
\includegraphics[width = 0.7\linewidth]{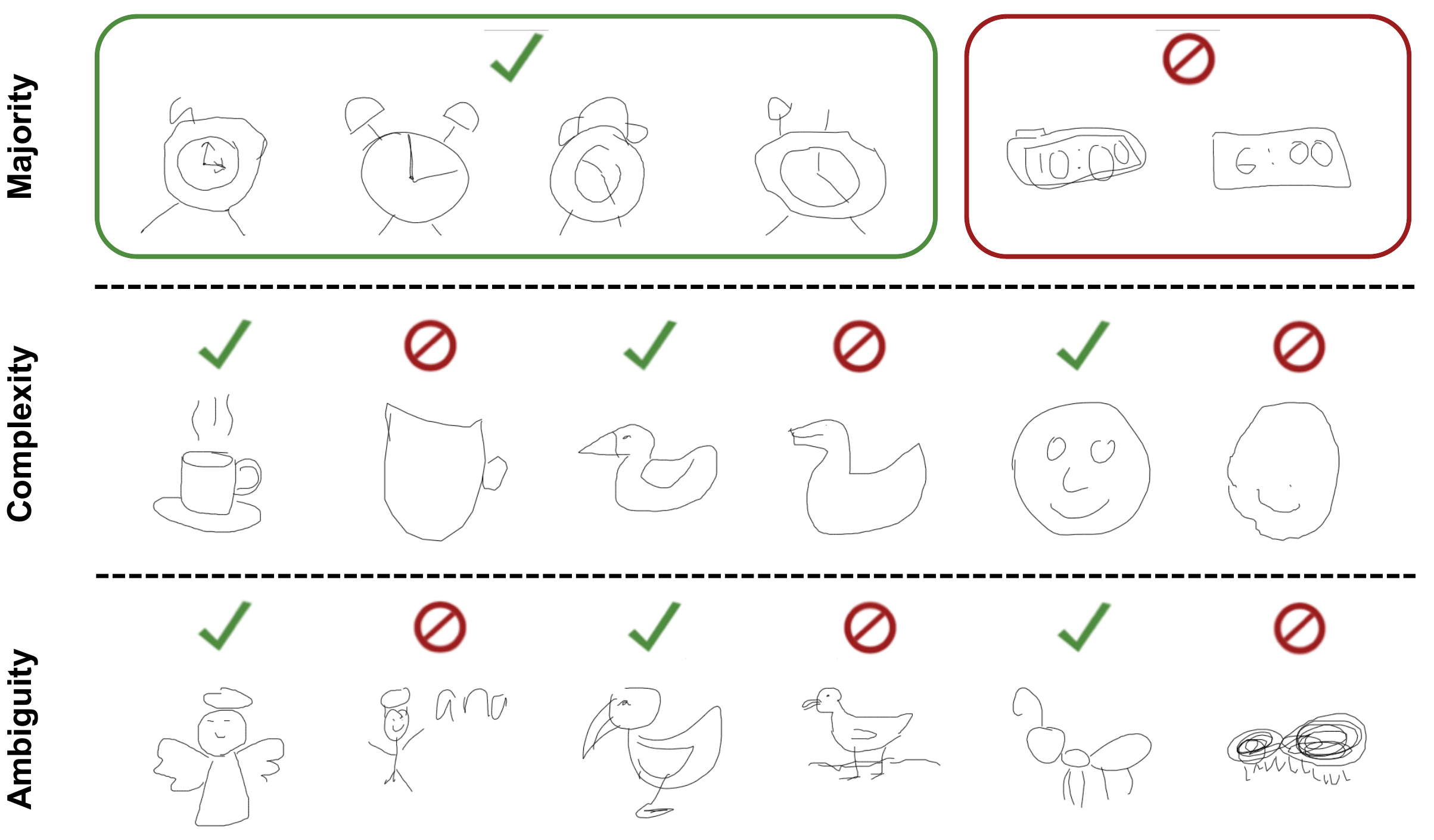}
\caption{Examples to illustrate our sketch selection process. See details in text.}
\label{fig:selection}
\end{figure}

\keypoint{Sketch Instance Selection:} Each QuickDraw category contains at least 100,000 sketches. Annotating all of them is not feasible. So 800 sketches are chosen from each category. First, some quality screening is performed. Specifically, since all QuickDraw sketches were drawn within 20 seconds, there are a large number of badly drawn sketches that are unrecognisable by humans, making part grouping impossible. We thus first discard sketches which could not be recognised by an off-the-shelf sketch classifier \cite{yu2017sketch}. The remaining sketches are then subject to the following instance selection criteria: (i) \textbf{Majority}: Sketches in each category can form subcategories which can be visually very different from each other. Only the sketches from the majority subcategory are selected. For example, the top row of Fig.~\ref{fig:selection} shows that most sketches from the alarm clock category belong to the ``with hand'' subcategory, whilst a small minority depicts digital clocks without hands. Only sketches from the former are selected. (ii) \textbf{Complexity:} Over-abstract sketches with less than three parts are removed. (iii) \textbf{Ambiguity:} Finally, we eliminate  sketches that contain both the target object and other objects/background to avoid ambiguity of the object category. Fig.~\ref{fig:selection} shows examples of how these criteria are enforced during instance selection.

\keypoint{Annotation:} Now given a sketch, each annotator is asked to group the strokes into groups. Each group has a semantic meaning and typically corresponds to an object part. So apart from the grouping label, the group ID is also annotated. Even though the group ID information is not used in our perceptual grouping model, it can be used when the task is sketch semantic segmentation. To obtain consistent grouping annotation, 25 annotators are recruited and each only annotates one category. Examples of the annotation can be seen in Fig.~\ref{annotation}. 

\section{Deep Universal Sketch Perceptual Grouper}
\subsection{Model Overview}
\begin{figure}[t]
\centering
\includegraphics[width = 0.7\linewidth]{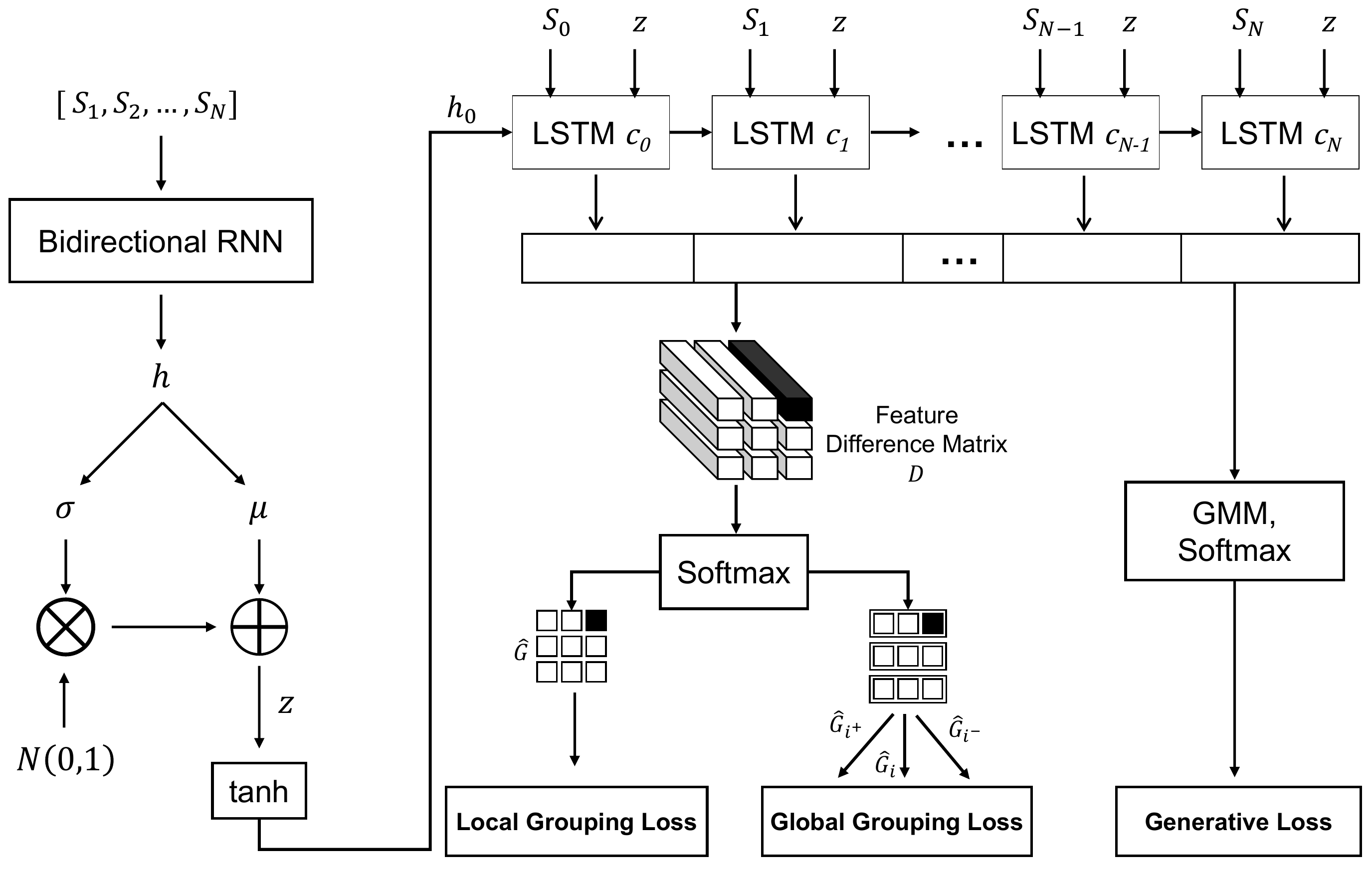}
\label{unseen}
\caption{A schematic of the proposed deep perceptual grouper. }
\label{fig:framework}
\end{figure}

Our deep sketch grouper is a variant of the sequence-to-sequence variational autoencoder (VAE) \cite{Bowman2015Generating,Kingma2013Auto}. As shown  in Fig.~\ref{fig:framework}, it is essentially a deep encoder-decoder with both the encoder and decoder being RNNs for modelling a sketch as a set of strokes. The encoder produces a global representation of the sketch, which is used as a condition for a variational decoder that aims to reconstruct the input sketch. Such sketch synthesis is a side task here. Our main aim is for the  decoder to produce a representation of each stroke useful for grouping them. Once learned, the decoder should implicitly model all the grouping principles used by the annotators in producing the grouping labels, so that the learned stroke representation can be used to compute a stroke affinity matrix indicating the correct stroke grouping.  To this end, the decoder has two branches: a generative branch to reconstruct the input sketch; and a discriminative branch that produces the discriminative stroke feature/affinity matrix.

\subsection{Encoder and Decoder Architecture}
Traditional perceptual grouping methods treat sketches as images composed of static pixels, thus neglecting the dependency between different segments and strokes (each stroke consists of a variable number of line segments). In our dataset, all the sketches are captured in a vectorised format, making sequential modelling of sketches possible. More specifically, we first represent a sketch as a sequence of \textcolor{black}{$N$} stroke-segments $[S_1, S_2, ..., S_N]$. Each segment is a tuple $(\Delta x, \Delta y, p)$, where $\Delta x$ and $\Delta y$ denote the offsets along the horizontal and vertical directions respectively, while $p$ represents the drawing state, following the same representation used for human handwriting \cite{graves2013generating}. 

With these stroke segments as inputs, both the encoder and decoder are RNNs. In particular, we adopt the same architecture as in sketch-rnn \cite{ha2017neural} for conditional sketch synthesis.  That is,  a bi-directional RNN \cite{Schuster1997Bidirectional} is used as the encoder to extract the global embedding of the input sketch. The final state output of the encoder is  then projected to a mean and a variance vector, to define an IID Gaussian distribution. That distribution is then sampled to produce a random vector $z$ as the representation of the input sketch. Thus $z$ is not a deterministic output of the encoder given a sketch, but a random vector conditional on the input. The decoder is an LSTM model. Its initial state  is conditional on $z$ via a single fully connected (FC) layer. At each time step, it then predicts the offset for each stroke segment in order to reconstruct the input sketch. For further details on the encoder/decoder architecture, please refer to  \cite{ha2017neural}.  

\subsection{Formulation}
\label{sec:objectives}

The decoder splits into two branches after the LSTM hidden cell outputs: a generative branch to synthesise a sketch and a discriminative branch for grouping. Different learning objectives are used for the two branches: in the generative branch, two losses encourage the model to reconstruct the input sketch; in the discriminative branch, the sketch grouping annotation is used to train the decoder to produce an accurate stroke affinity matrix for grouping. 

\keypoint{Group Affinity Matrix:}
The grouping annotation is represented as a sparse matrix denoting the group relationship between segments ${\bf{G}} \in \mathbb{R}^{N \times N}$. Denoting the $i^{th}$ sketch segment as $S_i,i \in [1,N]$, we have: 
\begin{equation}
\label{equ:adjacency}
G_{i,j}=\left\{
\begin{array}{rcl}
1, &  &if\  {S_i,S_j\ are\ from\ the\ same\ group}\\
0, &  &otherwise
\end{array} \right.
\end{equation}
Where each element of the matrix indicates whether the $i^{th}$ and $j^{th}$ segments belong to the same group or not.
A straightforward design of the discriminative learning objective is to make the affinity matrix computed using the learned stroke feature $f_i=\phi(S_i)$ as similar as possible to ${\bf{G}}$, via an $l_1$ or $l_2$ loss. However, we found that in practice this works very poorly. This is because  ${\bf{G}}$ conveys two types of grouping constraints: each element enforces a binary pairwise constraint for two segments, whilst the whole matrix also enforces global grouping constraint, e.g., if $S_1$ and $S_2$ are in the same group, and $S_2$ and $S_5$ are also in the same group, then global grouping consistency dictates that $S_1$ and $S_5$ must also belong to the same group. Balancing these two is critical because  pairwise grouping predictions are typically  noisy and can lead to global grouping inconsistency. However, using a single loss makes it impossible to achieve a balance. We thus propose to use two losses to implements the two constraints.

\keypoint{Local Grouping Loss:}
This loss requires that the pairwise relationship between two segments are kept when the pairwise affinity is measured using the learned stroke segment feature. The decoder LSTM learns a mapping function $\phi$ and map the $i^{th}$ stroke segment $S_i$ to a  128D feature vector $f_i$. To measure the affinity of any two segments in the input sketch, the absolute element-wise feature difference is computed to obtain a symmetric absolute feature difference matrix, ${\bf{D}}\in \mathbb{R}^{N \times N \times 128}$ as:
\begin{equation}
\label{equ:diff}
{\bf{D}} = \big\{D_{i, j} \  \big| \  i,j \in [1, N]\big\} = \big\{|f_{i} - f_{j}| \  \big| \  i,j \in [1, N]\big\}.
\end{equation}
Each vector $D_{i,j} \in \mathbb{R}^{128}$ is then subject to a binary classification loss (cross-entropy) to obtain the local affinity prediction $\hat{G}_{i,j}$, between the $i^{th}$ and $j^{th}$ segments. The local grouping loss, $\mathcal{L}_A$, is thus computed as:
\begin{equation}
\label{equ:aff_loss}
\mathcal{L}_A = \sum_{i=1}^{N}{\sum_{j=1}^{N}{\big[-G_{i,j}\log(\hat{G}_{i,j})-(1-G_{i,j})\log(1-\hat{G}_{i,j})\big]}}.
\end{equation}

\keypoint{Global Grouping Loss:} 
Using only a local grouping loss  may lead to global grouping inconsistency. However, formulating the global grouping consistency into a loss for a deep neural network is not straightforward. Our strategy is to first derive a global grouping representation for each segment using the local affinity prediction $\hat{G}_{i,j}$. We then use a triplet ranking loss to enforce that the segments belonging the same group have more similar grouping relationships to each other, than to a segment outside the group. Although the triplet ranking loss involves three segments only, since each segment is represented by its grouping relationship to all other segments, this loss is a global grouping loss.
More concretely, we first construct the local affinity prediction matrix ${\bf \hat{G}}$ with $\hat{G}_{i,j}$ as elements. Each row vector of ${\bf \hat{G}}$, $\hat{G}_{i,:}$ is then used as a global grouping relationship vector to represent $S_i$. The final global grouping loss $\mathcal{L}_G$ is:

\begin{equation}
\label{equ:group_loss}
\mathcal{L}_G=\max(0,\Delta+d(\hat{G}_{i,:},\hat{G}_{i^+,:})-d(\hat{G}_{i,:},\hat{G}_{i^-,:})),
\end{equation}
where ${i}$ represents an anchor segment, ${i^+}$ a positive segment in the same group and ${i^-}$ a negative segment from a different group, $\Delta$ is a margin and $d(\cdot)$ denotes a distance function between two feature inputs.  Here we take the squared Euclidean distance under the $l_2$ normalisation.

\keypoint{Generative Losses:} 
For the  generative branch, we use the same generative losses as in \cite{ha2017neural}. These include a reconstruction loss $\mathcal{L}_R$ and  a KL loss $\mathcal{L}_{KL}$ measuring the difference between the latent random vector $z$ and a IID Gaussian vector with zero-mean and unit variance. 

\keypoint{Full Learning Objective:}
Our full loss $\mathcal{L}_F$  can be written as:
\begin{equation}
\label{equ:full_loss}
\mathcal{L}_F = {\lambda_a}\mathcal{L}_A+{\lambda_g}\mathcal{L}_G+{\lambda_r}(\mathcal{L}_R+\mathcal{L}_{KL})
\end{equation}
where the hyper-parameters $\lambda_a$, $\lambda_g$, and $\lambda_r$ describe the relative importance of the different losses in the full training objective. 

\keypoint{Model Testing:}
During testing stage, given a sketch, the trained model is used to compute an estimated segment affinity matrix, $\hat{G}$. This affinity matrix is then used to generate the final grouping. Since the number of groups varies for different sketches, the group number also needs to be estimated. To this end, we adopt a recent agglomerative clustering method  \cite{Yang2016Joint} to obtain the final grouping. Note that the method does not introduce any additional free parameters. 

\subsection{Applications to Sketch Analysis}

\noindent {\bf Sketch Synthesis from Edge Map:} \quad A simple sketch synthesis method can be developed based on the proposed universal grouper. The method is based on grouping edge maps extracted from photo images and removing the least important groups. Assume that the $N$ segments of an edge map have been grouped in $K$ groups, denoted as $P_k, k \in [1,K]$. An importance measure is defined as:
\begin{equation}
\label{equ:imp_score}
I(P_k) = I_L(P_k) \cdot I_N(P_k) + I_D(P_k)
\end{equation}
where $I_L(P_k)$, $I_N(P_k)$ and $I_D(P_k)$ measure the importance from the perspectives of length, numbers and distribution of the segments in group $P_k$ respectively. A less important group has smaller number of segments with shorter lengths but occupies a bigger region. We thus have:
\begin{equation}
\label{equ:imp_len}
I_L(P_k) = \frac{\sum_{i=1}^{N_{P_k}}{L_{S_i}}}{\sum_{i=1}^N{L_{S_i}}},~
I_N(P_k) = \frac{N_{P_k}}{N},~
I_D(P_k) = \frac{\max(w, h){N_{P_k}}}{\sum_{i=1}^{N_{P_k}}{d(M_{P_k}, M_{S_i})}}
\end{equation}
where  $N_{P_k}$ is the number of segments in $P_k$, $L_{S_i}$ is the length of segment $S_i$, $w$ and $h$ are the width and height of the object, respectively, $M_{P_k}$ denotes the average position of group $P_k$ in the image plane, $M_{S_i}$ represents the average position of segment $S_i$, and Euclidean distance $d(\cdot)$ is used. With the importance measure $I(P_k)$ computed for each group, we can then drop the least important groups defined as those with  $I(P_k)<I_{\delta}$ where $I_{\delta}$ is a threshold. 

\keypoint{Fine-Grained Sketch Based Image Retrieval:} We further develop an unsupervised FG-SBIR method following \cite{Umar2018Abstraction}. Specifically, we apply our grouper to edge maps extracted from photos to synthesise human style sketches. Three threshold values of $I_{\delta}$ are used for each photo to accounts for the variable levels of abstraction among human sketchers. The photos and corresponding synthesised sketches are then used as data to train an off-the-shelf FG-SBIR model \cite{yu2016sketch}. During testing, the grouping and group removal processes are applied to the human sketches, again with three different thresholds. The matching scores using the three abstracted sketches plus the original query sketch are then fused to produce the final retrieval results. Note that for this unsupervised FG-SBIR model to work well, our grouper must be truly universal: it needs to work well on both human sketches which it was trained on, and photo edge maps. 

\section{Experiments}

\subsection{Datasets and Settings}
\keypoint{Dataset Splits and Preprocessing:}
Among the 25 categories in the new SPG dataset, we randomly select 20  as \textbf{seen categories}, and use the remaining 5 categories as \textbf{unseen categories} to test the generalisation of our universal grouper. In each seen category, we select 650 sketches for training, 50 for validation, and 100 for testing.  For the unseen categories, no data are used for training and we randomly select 100 sketches per category for testing to have the same per-category size as the seen categories. We normalise all the sketch strokes, and augment the sketch via stroke removal and distortion \cite{yu2017sketch}.

\keypoint{Implementation Details:}
Our deep grouper is implemented  on Tensorflow on a single Titan X GPU. For model training, we set the importance weights $\lambda_r$, $\lambda_a$ and $\lambda_g$ for different losses (Eq.~(\ref{equ:group_loss})) to 0.5, 0.6, and 1, respectively. The Adam optimiser \cite{kingma2014adam} is applied  with the parameters $\beta_1 = 0.5$, $\beta_2 = 0.9$, $\epsilon = 10^{-8}$. The initial  learning rate is set to 0.0003 with exponential weight decay. The model is trained for 22,000 iterations with a batch size of 100. 

\keypoint{Evaluation Metrics:}
Sketch perceptual grouping  shares many common characteristics  with the unsupervised image segmentation problem \cite{Xia2017W}. We thus adopt the  same metrics including  variation of information (VOI), probabilistic rand index (PRI), and segmentation covering (SC) as defined in \cite{Arbel2011Contour}. More detailed definition of these metrics in the context of sketch grouping are: (i) \textbf{Variation of Information:} In this metric, the distance between two groups in terms of their average conditional entropy is calculated. (ii) \textbf{Probabilistic Rand Index:} Rand index compares the compatibility of assignments between pairs of stroke segments in each group. (iii) \textbf{Segmentation Covering:} the overlapping between the machine grouping and human grouping is measured. For SC and PRI, higher scores are better, while for VOI, a lower score indicates better results. 


\keypoint{Competitors:}
Very few sketch perceptual grouping methods exist. The state-of-the-art model \textbf{Edge-PG} \cite{Qi2015Making} uses two  Gestalt principles, namely proximity (spatial closeness) and  continuity (slope trend) to compute an affinity matrix and  feeds the matrix to a  graph cut algorithm to get the groups. The weightings of the two principles are learned from data using RankSVM. This method thus differs from ours in that hand-crafted features are used and only two principles are modelled. Beyond sketch grouping, many semantic image segmentation methods have been proposed lately based on fully convolutional networks (FCN). We choose one of the state-of-the-art models,  \textbf{DeepLab} \cite{CP2016Deeplab} as a baseline. It is trained to take images as input and output the semantic grouping, i.e., each pixel is assigned a class label. A conditional random field (CRF) is integrated to the network to enforce the proximity and similarity principles. Note that: (1) DeepLab is a supervised semantic segmentation method. It thus needs not only grouping annotation as our model does, but also group ID annotation, which is not used by our model and Edge-PG. This gives it an unfair advantage. (2) It performs grouping at the pixel level whilst both our model and Edge-FG do it at the stroke/segment level. 



\subsection{Results on Perceptual Grouping}

\setlength{\tabcolsep}{8pt}
\begin{table}[tb]
\begin{center}
\resizebox{0.75\textwidth}{!}{
\begin{tabular}{lccc|ccc|ccc}
\toprule
\multirow{2}{*}{ \textbf{Category}}  & \multicolumn{3}{c|
}{\textbf{Ours}} & \multicolumn{3}{c|}{\textbf{Edege-PG \cite{Qi2015Making}}} & \multicolumn{3}{c}{\textbf{DeepLab \cite{CP2016Deeplab}}}\\
\noalign{\smallskip}
\cline{2-10} 
\noalign{\smallskip}
& \textbf{VOI $\downarrow$}  & \textbf{PRI $\uparrow$} & \textbf{SC $\uparrow$} & \textbf{VOI$\downarrow$ } & \textbf{PRI $\uparrow$} & \textbf{SC $\uparrow$} & \textbf{VOI $\downarrow$} & \textbf{PRI $\uparrow$} & \textbf{SC $\uparrow$}\\
\midrule
Airplane    & \textbf{0.58}   & \textbf{0.88}   & \textbf{0.78}   & 0.72   & 0.80   & 0.71   & 1.09   & 0.72   & 0.65   \\
Alarm clock & \textbf{0.46}   & \textbf{0.93}   & \textbf{0.83}   & 0.59   & 0.84   & 0.73   & 0.86   & 0.80   & 0.70   \\
Ambulance   & \textbf{0.67}   & \textbf{0.86}   & \textbf{0.77}   & 1.35   & 0.67   & 0.60   & 1.19   & 0.71   & 0.63   \\
Ant         & \textbf{0.86}   & \textbf{0.83}   & \textbf{0.69}   & 1.32   & 0.68   & 0.62   & 1.38   & 0.69   & 0.60   \\
Apple       & \textbf{0.25}   & \textbf{0.92}   & \textbf{0.91}   & 0.54   & 0.88   & 0.79   & 0.82   & 0.83   & 0.72   \\
Backpack    & \textbf{0.57}   & \textbf{0.88}   & \textbf{0.79}   & 1.29   & 0.70   & 0.61   & 1.59   & 0.67   & 0.59   \\
Basket      & \textbf{0.76}   & \textbf{0.84}   & \textbf{0.74}   & 1.27   & 0.71   & 0.59   & 1.37   & 0.69   & 0.61   \\
Butterfly   & \textbf{0.83}   & \textbf{0.76}   & \textbf{0.65}   & 1.30   & 0.69   & 0.58   & 1.58   & 0.66   & 0.58   \\
Cactus      & \textbf{0.51}   & \textbf{0.90}   & \textbf{0.83}   & 0.86   & 0.82   & 0.71   & 0.90   & 0.79   & 0.68   \\
Calculator  & \textbf{0.50}   & \textbf{0.86}   & \textbf{0.83}   & 0.98   & 0.77   & 0.68   & 1.17   & 0.72   & 0.64   \\
Camp fire   & \textbf{0.28}   & \textbf{0.95}   & \textbf{0.91}   & 1.05   & 0.71   & 0.65   & 0.77   & 0.85   & 0.74   \\
Candle      & \textbf{0.89}   & \textbf{0.78}   & \textbf{0.69}   & 1.47   & 0.65   & 0.57   & 1.54   & 0.67   & 0.60   \\
Coffee cup  & \textbf{0.38}   & \textbf{0.91}   & \textbf{0.86}   & 0.85   & 0.83   & 0.68   & 0.98   & 0.79   & 0.66   \\
Crab        & \textbf{0.69}   & \textbf{0.81}   & \textbf{0.74}   & 1.29   & 0.69   & 0.56   & 1.58   & 0.67   & 0.60   \\
Duck        & \textbf{0.86}   & \textbf{0.83}   & \textbf{0.69}   & 0.95   & 0.74   & 0.68   & 1.63   & 0.65   & 0.57   \\
Face        & 0.81   & \textbf{0.84}   & \textbf{0.74}   & 1.24   & 0.69   & 0.61   & \textbf{0.80}   & 0.82   & 0.73   \\
Ice-cream   & \textbf{0.41}   & \textbf{0.94}   & \textbf{0.85}   & 0.79   & 0.82   & 0.71   & 1.40   & 0.68   & 0.62   \\
Pig         & \textbf{0.63}   & \textbf{0.84}   & \textbf{0.78}   & 1.55   & 0.63   & 0.50   & 0.98   & 0.77   & 0.67   \\
Pineapple   & \textbf{0.50}   & \textbf{0.93}   & \textbf{0.82}   & 0.63   & 0.83   & 0.72   & 1.05   & 0.74   & 0.65   \\
Suitcase    & \textbf{0.54}   & \textbf{0.89}   & \textbf{0.83}   & 0.58   & 0.82   & 0.75   & 1.10   & 0.73   & 0.64   \\
\hline
Average     & \textbf{0.59}   & \textbf{0.87}   & \textbf{0.79}   & 1.03   & 0.75   & 0.65   & 1.20   & 0.73   & 0.65   \\
\bottomrule
\end{tabular}
}
\end{center}
\caption{Comparative grouping results on seen categories.}
\label{tab:mainresult}
\end{table}

\setlength{\tabcolsep}{10pt}
\begin{table}[tb]
\begin{center}
\resizebox{0.75\textwidth}{!}{
\begin{tabular}{lccc|ccc}
\toprule
\multirow{2}{*}{\textbf{Category}} 
& \multicolumn{3}{c|}{\textbf{Ours}} & \multicolumn{3}{c}{\textbf{Edge-PG \cite{Qi2015Making}}}\\
\cline{2-7} 
\noalign{\smallskip}
& \textbf{VOI $\downarrow$} &\textbf{PRI $\uparrow$} & \textbf{SC $\uparrow$} & \textbf{VOI $\downarrow$} & \textbf{PRI $\uparrow$} & \textbf{SC $\uparrow$}\\
\midrule
Angel       & \textbf{0.70}   & \textbf{0.87}   & \textbf{0.73}   & 1.19   & 0.69   & 0.60 \\
Bulldozer   & \textbf{0.81}   & \textbf{0.85}   & \textbf{0.73}   & 1.37   & 0.65   & 0.58 \\
Drill       & \textbf{0.67}   & \textbf{0.78}   & \textbf{0.77}   & 1.45   & 0.61   & 0.53 \\
Flower      & \textbf{0.39}   & \textbf{0.90}   & \textbf{0.84}   & 0.79   & 0.75   & 0.64 \\
House       & \textbf{0.46}   & \textbf{0.91}   & \textbf{0.83}   & 0.85   & 0.77   & 0.69 \\
\hline
Average     & \textbf{0.64}   & \textbf{0.86}   & \textbf{0.77}   & 1.13   & 0.69   & 0.61 \\
\bottomrule
\end{tabular}
}
\end{center}
\caption{Perceptual grouping results  on unseen categories.}
\label{tab:unseen}
\end{table}

\begin{figure}[tb]
\centering
\includegraphics[width = 0.7\linewidth]{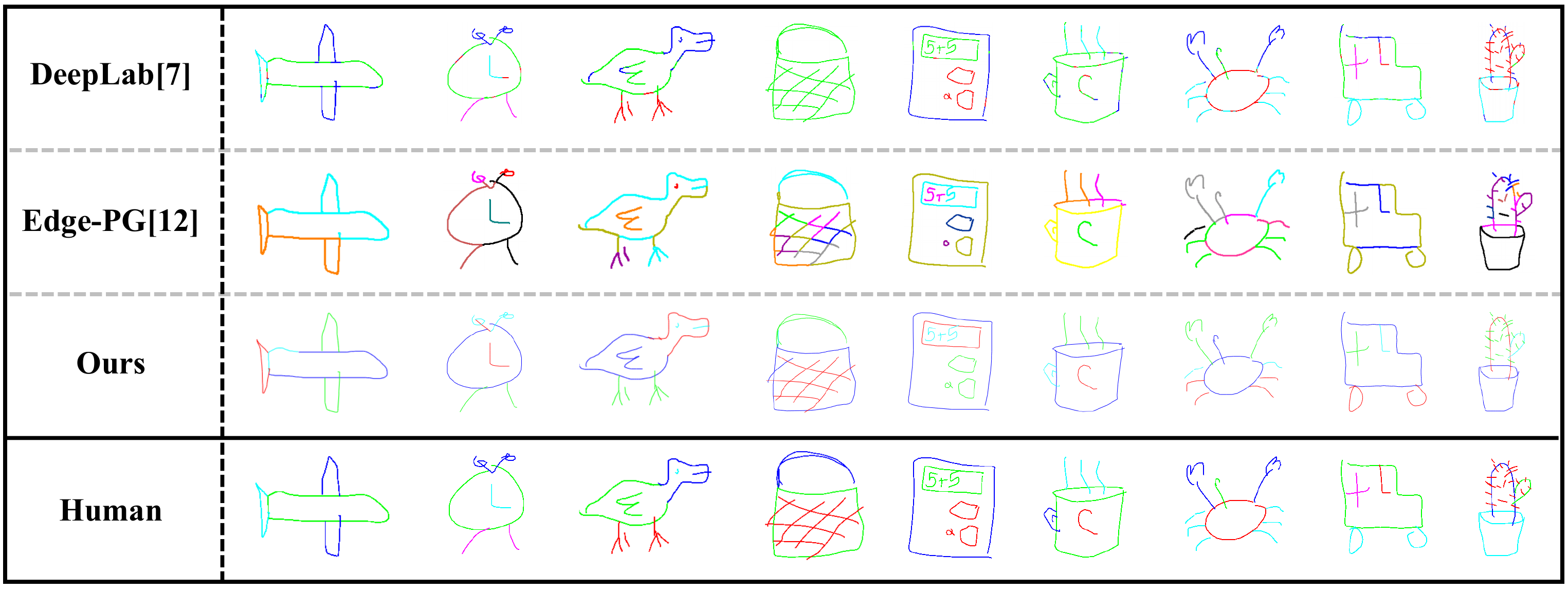}
\caption{Qualitative grouping results on seen categories.}
\label{fig:vis_seen}
\end{figure}

\keypoint{Results on Seen Categories:}
In this experiment, the model are trained on the seen category training set and tested on the seen category testing set.   From Table~\ref{tab:mainresult}, we can see that: (i) Our model achieves the best performance across all 25 categories on each metric, except the VOI metric on face where our model is slightly inferior to DeepLab. The improvement on VOI is particularly striking indicating that the groups discovered by our model in each sketch are distinctive to each other. In contrast, the two compared models tend to split a semantic part into multiple groups resulting similar groups (see Fig.~\ref{fig:vis_seen}). (ii) Edge-PG is much worse than our method because it is based on hand-crafted features for only two principles, while our model implicitly learns the features and combination strategy based on end-to-end learning from human group annotation.
(iii) Although DeepLab also employs a deep neural network and uses additional annotations, its result is no better than Edge-PG. This suggests that for sketch perceptual grouping, it is important to  treat sketches as a set of strokes rather than pixels, as strokes already grouping pixels. These constraints are ignored by the DeepLab types of models designed for photo image segmentation. 

\begin{figure}[tb]
\centering
\includegraphics[width = 0.7\linewidth]{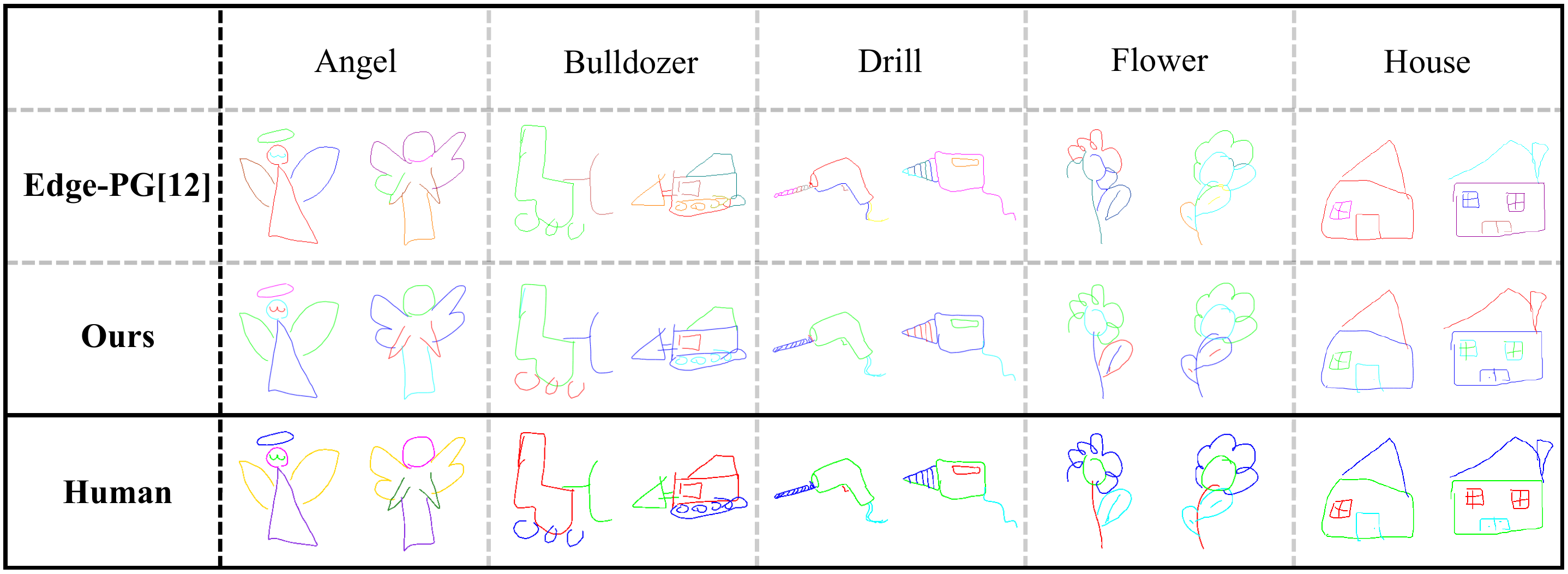}
\label{seen}
\caption{Qualitative grouping results on unseen categories.}
\label{fig:vis_unseen}
\end{figure}

Some examples of the grouping results are shown in Fig.~\ref{fig:vis_seen}. As expected, ignoring the stroke level grouping constraint on pixels, each stroke is often split into multiple groups by DeepLab \cite{CP2016Deeplab}. Edge-PG \cite{Qi2015Making} does not suffer from that problem. However, it suffers from the limitations on modelling only two principles. For example, to group the clock contour (second column) into one group, the closure principle needs to be used. It is also unable to model even the two principles effectively due to the limited expressive power of hand-crafted features: in the airplane example (first column), the two wings should be grouped together using the continuity principle, but broken into two by Edge-PG.  In contrast, our model produces more consistent groupings using multiple principles dynamically. For instance, both DeepLab and our model successfully deploy the similarity principle to group the two legs of both the alarm clock (second column) and duck (third column) together. But DeepLab does so by explicitly encoding the principle in its CRF layer, while our model does it implicitly. In the cactus example (last column), to produce the correct grouping of those spikes, both continuity, similarity and less prevalent  principles such as common fate need to be combined. Only our model is able to do that because it has implicitly learned to model all the principles used by humans to annotate the groupings.

\keypoint{Results on Unseen Categories:}
In this experiment,  models learned using seen categories are tested directly on  unseen categories without any  fine-tuning. This experiment is thus designed to evaluate whether the grouper is indeed universal, i.e., can be applied to any new object category. Note that as a supervised segmentation method, DeepLab cannot be applied here because each category has a unique set of group IDs. The results of our model and Edge-GP are shown in Table~\ref{tab:unseen}. It can been seen that our model outperforms Edge-GP by a big margin. Importantly, comparing Table \ref{tab:unseen} with Table~\ref{tab:mainresult}, our model's performance on PRI and SC hardly changed. In contrast, the Edge-PG's performance on the unseen categories is clearly worse than that on the seen categories. This suggests that our grouper is more generalisable and universal. Some qualitative results are shown in Fig.~\ref{fig:vis_unseen}. It again shows that the lack of powerful feature learning and limitation on only two principles contribute to the weaker results of Edge-GP. 


\begin{table}[t]
\begin{center}
\begin{tabular}{lccc}
\hline
\textbf{Method}  & \textbf{VOI $\downarrow$} & \textbf{PRI $\uparrow$} & \textbf{SC $\uparrow$}  \\
\hline
Edge-PG \cite{Qi2015Making}      & 1.69          & 0.62          & 0.53 \\
Ours                             & \textbf{0.96} & \textbf{0.78} & \textbf{0.71}\\
\hline
\end{tabular}
\end{center}
\caption{Compare Edge-PG \cite{Qi2015Making} with ours  on Edge-PG \cite{Qi2015Making}'s datasets.}
\label{tab:segmentation}
\end{table}

\setlength{\tabcolsep}{10pt}
\begin{table}[tb]
\begin{center}
\resizebox{0.75\textwidth}{!}{
\begin{tabular}{lccc|ccc}
\toprule
\multirow{2}{*}{\textbf{Method}} 
& \multicolumn{3}{c|}{\textbf{Seen Categories}} & \multicolumn{3}{c}{\textbf{Unseen Categories}}\\
\cline{2-7} 
\noalign{\smallskip}
& \textbf{VOI $\downarrow$} &\textbf{PRI $\uparrow$} & \textbf{SC $\uparrow$} & \textbf{VOI $\downarrow$} & \textbf{PRI $\uparrow$} & \textbf{SC $\uparrow$}\\
\midrule
Ours - A - G      & 1.45         & 0.65         & 0.59         &1.53           &0.64          &0.56\\
Ours - R - G      & 1.12         & 0.71         & 0.64         &1.36           &0.68          &0.59\\
Ours - R - A      & 1.27         & 0.69         & 0.63         &1.48           &0.64          &0.57\\
Ours - G          & 0.63         & 0.86         & 0.78         &0.71           &0.84          &0.73\\
Ours - A          & 0.75         & 0.80         & 0.72         &0.95           &0.78          &0.67\\
Ours - R          & 0.68         & 0.83         & 0.76         &0.86           &0.78          &0.69\\
Ours + $l_2$      & 2.68         & 0.58         & 0.49         &2.63           &0.59          &0.49\\
\hline
Ours full model   &\textbf{0.59} &\textbf{0.87} &\textbf{0.79} &\textbf{0.64}  &\textbf{0.86} &\textbf{0.77} \\
\bottomrule
\end{tabular}
}
\end{center}
\caption{\textcolor{black}{Performance of different variants of our model on seen and unseen categories.}}
\label{tab:gap_unseen}
\end{table}


\begin{table}[t]
\begin{center}
\resizebox{0.8\textwidth}{!}{
\begin{tabular}{lcc|cc}
\hline
\multirow{2}{*}{\textbf{Method}} & \multicolumn{2}{c}{\textbf{Shoe-V2}} & \multicolumn{2}{c}{\textbf{Chair-V2}}\\
\cline{2-5}
 & \textbf{Top1} & \textbf{Top10} & \textbf{Top1} & \textbf{Top10}\\
\hline
Scribbler \cite{Sangkloy2017Scribbler} & 8.86\%             & 32.28\%            & 31.27\%            & 78.02\% \\
LDSA \cite{Umar2018Abstraction}        & 21.17\%            & 55.86\%            & 41.80\%            & 84.21\% \\
Ours                                   & \textbf{26.88\%}   & \textbf{61.86\%}   & \textbf{45.57\%}   & \textbf{88.61\%} \\
\hline
Upper Bound                            & 34.38\%            & 79.43\%            & 48.92\%            & 90.71\% \\
\hline
\end{tabular}
}
\end{center}
\caption{FG-SBIR performance on Shoe-V2 and Chair-V2 dataset}
\label{tab:fgsbir}
\end{table}

\keypoint{Results on Unseen Dataset:}
To further demonstrate the generalisation ability of our universal grouper, we test the trained model on a different dataset. Specifically, we choose 10 categories from the dataset in \cite{Qi2015Im2Sketch} including 5  categories overlapping with our dataset and 5 new categories. Note that the sketches in this datasets are from the  database proposed in \cite{Eitz2012How}, which are drawn without the 20 second constraint, thus exhibiting much more details with better quality in general. This dataset thus represents a different domain.  Table \ref{tab:segmentation} shows that our model again demonstrates better generalisation ability than Edge-PG.

\keypoint{Ablation Study:} Our model is trained with a combination of generative and discriminative losses (Sec.~\ref{sec:objectives}). These include the local grouping loss $\mathcal{L}_A$, global grouping loss $\mathcal{L}_G$, reconstruction loss $\mathcal{L}_R$ and KL loss $\mathcal{L}_{KL}$. Among them, all but the KL loss can be removed, leading to six variants of our model, e.g., Ours - A - G is obtained by removing $\mathcal{L}_A$ and $\mathcal{L}_G$. In addition, we implement Ours + $l_2$ which is having an $l_2$ loss on the predicted affinity matrix $\hat{G}$ w.r.t. the ground truth matrix $G$ to examine the importance of having separate local and global grouping losses. The results are shown in Table~\ref{tab:gap_unseen}. Clearly all three losses contribute to the performance of our model. The poorest result was obtained when an $l_2$ loss is added on the predicted affinity matrix, suggesting that balancing the local and global grouping losses is critical for learning a good grouper. We further show that the improvement of our full model over Ours - R on the unseen categories (0.64 vs 0.86) is bigger on the seen categories (0.59 vs. 0.68). This indicates that the generative loss helps the model generalise to unseen categories. 

\begin{figure}[tb]
\centering
\includegraphics[width = 0.85\linewidth]{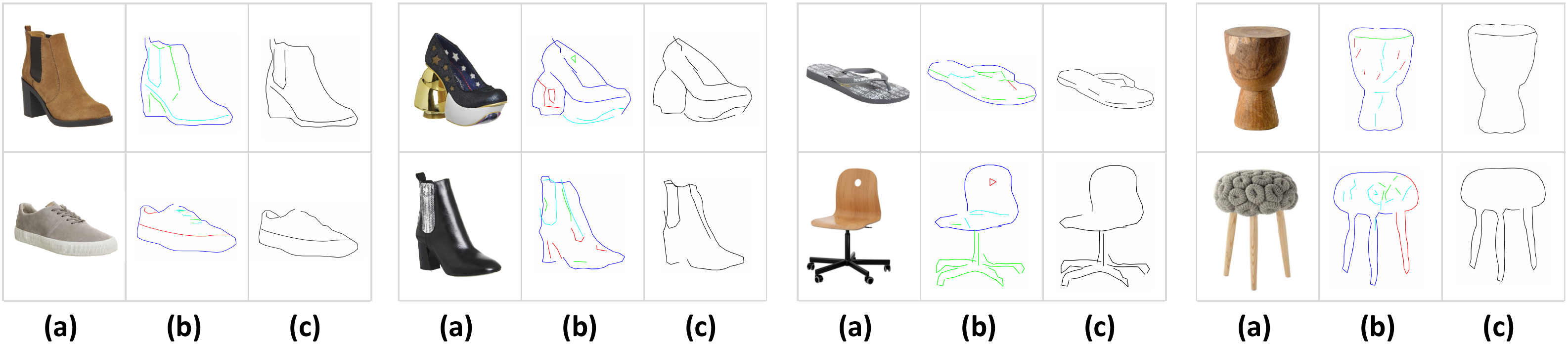}
\caption{Applying our grouper to synthesise abstract sketches from photo edge map. \textbf{(a)} columns show the photos; \textbf{(b)} columns give the edge maps extracted from  the photos and the grouping results;  \textbf{(c)} columns provide synthesised abstract sketches.}
\label{fig:fgsbir}
\end{figure}

\subsection{Applications on Sketch Synthesis and FG-SBIR}
One application of our grouper is to use it as an abstraction model so that edge maps extracted from photos can be grouped and abstracted to synthesise human-like sketches. Fig.~\ref{fig:fgsbir} shows some examples of  edge map grouping results and the synthesised sketches. It can be seen that our grouper is generalisable to photo edges and our abstraction method produces visually appealing sketches. The synthesised sketches are then used to train a state-of-the-art FG-SBIR model \cite{yu2016sketch} without using any real human sketches. We use the largest FG-SBIR datasets QMUL Shoe-V2 and Chair-V2 \cite{sketchx}. \textcolor{black}{We first compare with the same FG-SBIR model trained using synthesised sketches from the deep conditional GAN network in \cite{Sangkloy2017Scribbler} (denoted as Scribbler). As can be seen in Table~\ref{tab:fgsbir}, our model performs much better. This suggests that our edge abstraction model, albeit simple, synthesises more realistic sketches from edge maps. We further} compare with a recently proposed unsupervised FG-SBIR model LDSA \cite{Umar2018Abstraction} which is also based on abstracting photo edge maps to synthesise sketches. Table \ref{tab:fgsbir} shows that our model outperforms LDSA model by 5.71\% and 3.63\% on top 1 accuracy on Shoe-V2 and Chair-V2, respectively. The results are not far off the upper-bound which is obtained using the same FG-SBIR model trained with the real sketch-photo pairs in Shoe-V2 and Chair-V2. This shows that our method enables FG-SBIR to be used  without the expensive collection of sketch-photo pairs.

\section{Conclusion}

We have proposed an end-to-end sketch perceptual grouping model. This is made possible by collecting a new large-scale sketch grouping dataset SPG. Our grouper is trained with generative losses to make it generalisable to new object categories and datasets/domains. Two grouping losses were also formulated to balance the local and global grouping constraints. Extensive experiments showed that our model significantly outperforms existing groupers. We also demonstrated our grouper's application to sketch synthesis and FG-SBIR.

\bibliographystyle{splncs}
\bibliography{2836}
\end{document}